\documentclass[lettersize,journal]{IEEEtran}
\usepackage{amsmath,amsfonts}
\usepackage{array}
\usepackage{textcomp}
\usepackage{stfloats}
\usepackage{url}
\usepackage{verbatim}
\usepackage{graphicx}
\usepackage{cite}

\usepackage[utf8]{inputenc} 
\usepackage[T1]{fontenc}    
\usepackage{hyperref}       
\usepackage{url}            
\usepackage{booktabs}       
\usepackage{amsfonts}       
\usepackage{nicefrac}       
\usepackage{microtype}      
\usepackage{lipsum}
\usepackage{algorithm}
\usepackage{algorithmic}
\usepackage{fancyhdr}       
\usepackage{graphicx}       
\graphicspath{{media/}}     
\usepackage{amsmath}
\usepackage{subfigure}
\usepackage{balance}

\begin{document}

\title{Human-like Decision-making at Unsignalized Intersection using Social Value Orientation}

\author{Yan Tong$^{1,2}$, Licheng Wen$^{1}$, Pinlong Cai$^{1, \ast}$, Daocheng Fu$^{1}$, Song Mao$^{1}$, Yikang Li$^{1}$
\thanks{$^{1}$ Shanghai Artificial Intelligence Laboratory, Shanghai, China.}
\thanks{$^{2}$ College of Automotive Engineering, Jilin University, Changchun, China.}
\thanks{$\ast$ Corresponding author.}
}


\maketitle

\begin{abstract}
With the commercial application of automated vehicles (AVs), the sharing of roads between AVs and human-driven vehicles (HVs) becomes a common occurrence in the future. While research has focused on improving the safety and reliability of autonomous driving, it's also crucial to consider collaboration between AVs and HVs. Human-like interaction is a required capability for AVs, especially at common unsignalized intersections, as human drivers of HVs expect to maintain their driving habits for inter-vehicle interactions. This paper uses the social value orientation (SVO) in the decision-making of vehicles to describe the social interaction among multiple vehicles. Specifically, we define the quantitative calculation of the conflict-involved SVO at unsignalized intersections to enhance decision-making based on the reinforcement learning method. We use naturalistic driving scenarios with highly interactive motions for performance evaluation of the proposed method. Experimental results show that SVO is more effective in characterizing inter-vehicle interactions than conventional motion state parameters like velocity, and the proposed method can accurately reproduce naturalistic driving trajectories compared to behavior cloning.
\end{abstract}


\section{Introduction}
The development of automated vehicles (AVs) is a hot topic of global academic and industrial concern. Many commercial companies have released their vehicles with different levels of automation. The Waymo One service allows passengers to call and use Robotaxis  for transportation through mobile applications \cite{lebeau2018waymo}. Tesla has launched the Autopilot system, which can provide drivers with advanced safety and convenience functions such as adaptive cruise control and lane-keeping assistance \cite{ingle2016tesla}. Apollo Go is a Robotaxi launched by Baidu, which is equipped with automated driving technology and has already conducted experimental operations in cities such as Beijing and Changsha \cite{Daws2020baidu}. Although many existing research focuses on how to develop automated technologies to improve the safety and reliability of vehicles \cite{wang2021review, cai2022general}, the application coverage of these advanced technologies on vehicles is still relatively low. Due to technological and economic reasons, it will take a long time for AVs to completely replace traditional human-driving vehicles (HVs). Therefore, sharing roads between AVs and HVs will be a frequent occurrence in the future \cite{chen2020future, shi2022empirical}.

The collaboration between AVs and HVs has a significant impact on traffic flows, which has led many researchers to attempt to construct interaction models for AVs. Game theory \cite{li2020game, li2023human}, graph-based methods \cite{deng2020conflict, chen2022conflict}, reinforcement learning (RL) \cite{li2020deep, bai2022hybrid}, and other methodologies have been introduced to model the inter-vehicle interaction. Although AVs can have more intelligent interaction methods, it is difficult for AVs to understand the intention of human drivers when they meet HVs on the roads \cite{alambeigi2020crash, wang2020decision}. Unconventional interaction methods for AVs may confuse human drivers, which not only affects the efficiency of inter-vehicle interaction, but also leads to the exclusion of AVs by the public. Therefore, human-like interaction is a necessary capability for AVs \cite{wu2023deep}, as human drivers hope to maintain their past practice for inter-vehicle interaction. 

Social value orientation (SVO) haves been proven to be feasible in describing different types of inter-vehicle interactions such as car-following \cite{wen2022modeling}, lane-changing \cite{zhao2021yield}, and ramp-merging \cite{schwarting2019social, wang2021social} motions. 
However, the interaction between vehicles at unsignalized intersections is much more complex than the above motions. Few works have studied int the interaction behavior of vehicles at unsignalized intersections \cite{trentin2023multi, tian2020game}, but have not taken into account human psychological factors that affect decision-making in these studies.

Therefore, we proposed an SVO-guided reinforcement learning (RL) method by considering both interactive interpretability and model generalization. First, we devise a quantitative calculation method for conflict-related SVOs to describe the inter-vehicle interactions between vehicles from different approaches at unsignalized intersections. Next, we utilize the SVO as part of the reward functions to enhance the RL model for the decision-making of vehicle motion. Finally, we evaluate the performance of the proposed method on a naturalistic driving dataset with many highly interactive scenarios. The contributions of this paper can be summarized as following.
\begin{itemize}
    \item We demonstrate that SVO has significant advantages in characterizing vehicle interactions at unsignalized intersections compared to conventional motion state parameters like velocity. The proposed quantification approach of SVO can describe that the vehicle interaction process is determined by the trade-off between altruism and individualism to determine immediate actions.
    \item We establish a human-like interactive decision-making model to guide vehicles to cross the unsignalized intersections, including straight-going, left-turning, and right-turning motions. Under the guidance of SVO and other reward functions, the RL-based method can simulate the behaviors of human drivers, completing observation and judgment, and output sequential actions.
    \item We reconstruct near-real vehicle interaction scenarios based on the proposed method. Compared to the behavior cloning and vanilla RL-based method, the characteristics of the planned trajectories by the proposed method can be closer to the trajectories of HVs in naturalistic driving scenarios.
\end{itemize}

\section{Related work}
\subsection{Trajectory Planning at Unsignalized Intersection}
The trajectory planning for vehicles in unsignalized intersections is a challenging problem that has garnered much interest. To achieve reasonable trajectory planning and avoid conflicts, drivers must observe the movements of vehicles from different approaches. However, the road conditions at unsignalized intersections are complex, making driving tasks highly demanding. With the advent of AVs, researchers are interested in new ways of intersection management. Dresner and Stone proposed an autonomous intersection management method based on a reservation system, by which AVs could receive passage permits from the control center provided that the spatiotemporal conflicts of traffic flow in different directions are avoided \cite{dresner2004multiagent}. Wu et al. optimized the passing order of AVs by a Markov decision-making method for the multi-agent systems, which reduced vehicle delays without collision constraints \cite{wu2019dcl}. Chen et al. proposed an autonomous management strategy using the deep RL method to enhance the crossing efficiency of vehicles at intersections, which significantly improved over the first-come, first-served strategy \cite{chen2019intersection}. 

Although plenty of autonomous management methods have been applicable to AVs at unsignalized intersections, new strategies need to be explored for current intersections that are shared by AVs and HVs. Naidja et al. presented a game-theoretic trajectory planner and decision-maker for mixed-traffic environments, which used a clothoid interpolation method to generate human-like trajectories and reduced the dimension of the search space for trajectory optimization \cite{naidja2023interactive}. 
Zhou et al. proposed an unsignalized intersection management strategy for connected AVs and HVs that included a heuristic priority-queue-based right-of-way allocation algorithm and a vehicle planning and control algorithm \cite{zhou2022unsignalized}. Yan et al. proposed and implemented a novel approach to optimize traffic flow at unsignalized intersections in mixed traffic situations using a deep RL method, which significantly improved traffic flow through unsignalized intersections and provided better performance compared to many existing methods \cite{yan2021courteous}. Li et al. proposed a game-theoretic decision algorithm that considers social compatibility, which was found to improve both safety and time efficiency in human-in-the-loop experiments involving 24 subjects interacting with AVs at an unsignalized intersection, based on the analysis of 207 intersection interaction cases from 4300 video clips of traffic accidents \cite{li2023safe}. However, intelligent but non-human decision-making often overlooks the social interaction between vehicles, making it difficult for these methods to be applied in practice. 

\subsection{Social Interaction for Traffic Participants}
In recent years, the social interaction among traffic participants has gradually attracted the interest of researchers. Helbing and Molnár proposed a social force model to explain inter-pedestrian interaction movements through virtual attraction and repulsion forces \cite{helbing1995social}, which was later extended to the interaction between pedestrians, vehicles, and roads \cite{yang2018social}. However, transient social forces cannot describe the sequential trajectories of the interaction process and be incompatible with structured road and vehicle kinematics. Lu et al. proposed a novel car-following and control model based on the risk homeostasis theory \cite{lu2012quantitative, zhang2019impact}. Wang et al. proposed a driving safety field model to describe the driver-vehicle-road interactions, which provided a physical explanation of the physical spatial relationships and motion states in dynamic traffic scenarios \cite{wang2015driving, wang2016driving}. However, parameter calibration for such a driving safety field is very difficult, and the superposition of coupling risk in complex multi-vehicle scenarios is hard to handle. 

Neural network-based methods are currently prevalent because researchers believe that complex social interactions are difficult to model through mechanistic models and rely on data-driven strong generalization capabilities. Sun et al. decoupled the joint prediction problem of multi-vehicle interaction into a marginal prediction problem, outputting decisions of passing and yielding \cite{sun2022m2i}. Then, a tree structure based on vehicle interaction relationships was constructed to avoid trajectory interaction collisions, and applied to scenario editing of naturalistic driving datasets \cite {sun2022intersim}. Zhou et al. divided continuous scenarios into multiple local regions for implicit representation of spatiotemporal interaction relationships, and achieved information transfer in local regions to approximate global interaction \cite{zhou2022hivt}. Besides, graph-based approaches are used to construct the nodes and edges of a directed graph from the road environment, vehicles, and their interrelationships in the scenarios, and fed them into a deep neural network to analyze the implicit features of the interaction \cite{li2019grip, sheng2022graph}. Although learning-based methods can achieve good performance in open-source datasets, they require more exploration in terms of interpretability.

\subsection{The Concept and Application of the SVO}
The development of behavioral and cognitive science brings innovative thinking to the human-involved decision-making process. Social behavior can be explained by the characteristics of altruism and individualism, which jointly affect the behavioral intentions of drivers on the road. The SVO is a typical interpersonal trait, which is the social preference exhibited by individuals in resource allocation or risk decision-making for themselves and others \cite{van1991social}. Some researchers considered the SVO as a significant variable in understanding multi-vehicle interaction and decision-making processes, and proposed interaction-aware methods by quantifying altruism and individualism.
Schwart et al. proposed a framework that integrated social psychology tools into the controller design of AVs, and predicted the human drivers' behaviors by estimating their SVOs, which was verified in the cases of merging and unprotected left-turning motions on highways \cite{schwarting2019social}. Wang et al. proposed an interactive perception security assessment framework for AVs and applied it to the roundabout entering scenarios, in which the level-k game theory and SVO were used to characterize the interaction behavior \cite{wang2022comprehensive}.  Crosato et al. integrated the SVO into the RL framework to adjust the behaviors of AVs towards pedestrians from reckless to cautious \cite{crosato2021human}. Buckman et al. proposed a coordinated strategy that managed AVs at intersections based on the SVOs of vehicles, thereby reducing the average delay of vehicles \cite{buckman2019sharing}. Zhao et al. proposed an active semantic decision-making method to deal with the parallel interaction between HVs and AVs through a game theory model based on quantitative social preference and counterfactual reasoning \cite{zhao2021yield}.

\section{Methodology}
In this section, we propose a human-like decision-making model considering the social interaction between vehicles for determining vehicle trajectories at the unsignalized intersection. Firstly, we provide a description of the concerned problem to be solved. Then, the reward function combined with the social interaction is designed. Finally, the designed reward function is employed to enhance the RL method, and a detailed implementation process for the proposed method is provided.

\subsection{Problem Description}
\begin{figure*}
    \centering
    \includegraphics[width=4.5in]{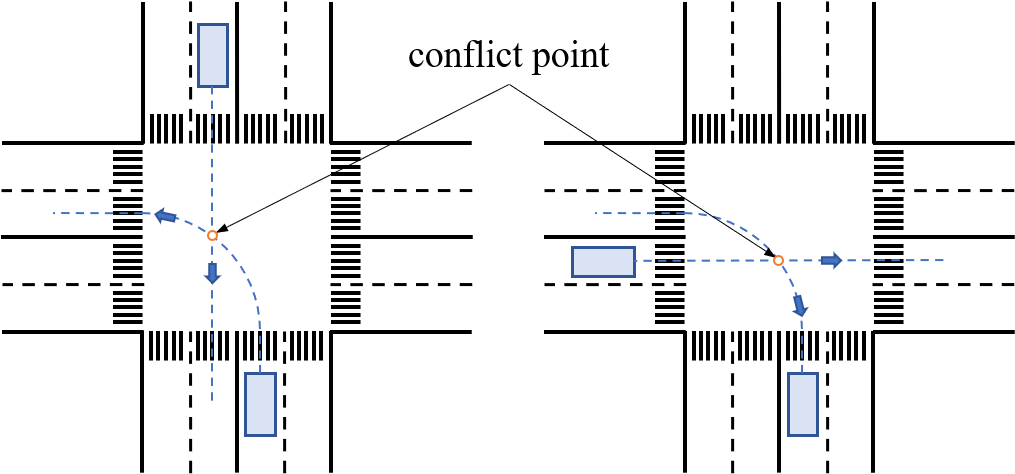}
    \caption{Interaction Scenario.}
    \label{fig:Problem Description}
\end{figure*}

Consider the interaction behavior of two vehicles at an unsignalized intersection, where each vehicle observes the state $s_t$ at time step $t$ and outputs the corresponding action $a_t$ according to its state. The initial states $s_0$ of the vehicles are taken directly from the naturalistic driving dataset. Since the crucial factor for vehicle interaction within an intersection is longitudinal displacement, this paper considers only the control of vehicle longitudinal motion for simplicity. Thus, we assume that vehicles always follow the given routes, which are also provided by the naturalistic driving dataset. The conflict points are defined as the positions in the concerned unsignalized intersection scenarios where potential collisions between vehicles may occur. Since the vehicle positions are not time-aligned, the conflict points for two directed vehicle trajectory curves can be regarded as the first points in two curves where the distances to the intersection point are less than a safety threshold (5 m in this paper). The general interaction scenarios at unsignalized intersections are shown in Fig. \ref{fig:Problem Description}.

Vehicle state can be represented by several motion parameters. 
We choose the parameters of position, velocity, and heading information that are closely related to the inter-vehicle interaction process. The multi-vehicle interaction process can always be split into pairwise interactions for ease of handling. A two-row vector is used to describe the kinematic state of the ego vehicle and the other vehicle by:
\begin{equation}
    s_t = 
    \begin{bmatrix}
        x_{t} & y_{t} & v_{t} & \varphi_{t}\\
        \hat{x}_{t}  & \hat{y}_{t} & \hat{v}_{t}  & \hat{\varphi}_{t}
    \end{bmatrix},
\end{equation}
where $(x_{t}, y_{t})$, $v_{t}$, and $\varphi_{t}$ are the position, velocity, and direction of the ego vehicle, respectively. $(\hat{x}_{t}, \hat{y}_{t})$, $v_{t}$, $\hat{\varphi}_{obs}$ are the position, velocity, and direction of the other vehicle, respectively. We study the social interaction behavior of vehicles in terms of velocity, and the control of longitudinal velocity better reflects the drivers' scrambling and avoidance during regular driving. Given the driving trajectory, we take the longitudinal acceleration $[a_{t},\hat{a}_{t}]$ of the self and other vehicles as continuous decision actions.

\subsection{Reward Function Considering Social Interaction}

\subsubsection{Social Interaction Modeling}
In this paper, we want to learn the strategies of interaction behavior from naturalistic driving data. Social interaction modeling is pivotal to the decision-making process of vehicle behavior. Social value can be employed to measure the outcomes of self and others. For the intersection scenario, velocity is an important parameter that directly responds to vehicle outcome, which reflects a series of driving characteristics and behaviors such as the efficiency of passage and right-of-way contention. Moreover, the distance between the vehicle and the conflict point has a corresponding connection to the driver's judgment of the intensity of the interaction. Since the interaction is a process involving both parties, when the other vehicle is closer to the conflict point, the ego vehicle will consider its own and others' outcome allocation more, and the outcome involved with social value will increase accordingly. According to the above analysis, we define the outcome of both sides of the interaction as:
\begin{equation}
    U=v/d,\ \hat{U}=\hat{v}/\hat{d},
\end{equation}
where $U$ and $\hat{U}$ denote the outcome of self and the other. $v$ and $\hat{v}$ denote the velocity of self and the other. $d$ and $\hat{d}$ are the distances of self and the other to the conflict point.

We use the SVO to characterize the social interaction behavior among vehicles. As a terminology in computational psychology, the SVO describes the individual's allocation of outcomes to self and outcomes to the other, which is used to reflect an individual's social preference. Based on this theory, the utility of an individual can be expressed as a linear combination of outcomes to self and other people\cite{mcclintock1978social}. More generally, social interaction can be expressed by the trigonometric function \cite{liebrand1988ring}:
\begin{equation}
    U_f=U\cos(\varphi)+\hat{U}\sin(\varphi),   
\end{equation}
where $\varphi$ is the SVO of the ego vehicle and $U_f$ is the final utility of the ego vehicle.

\begin{figure}
    \centering
    \includegraphics[width=6.5cm]{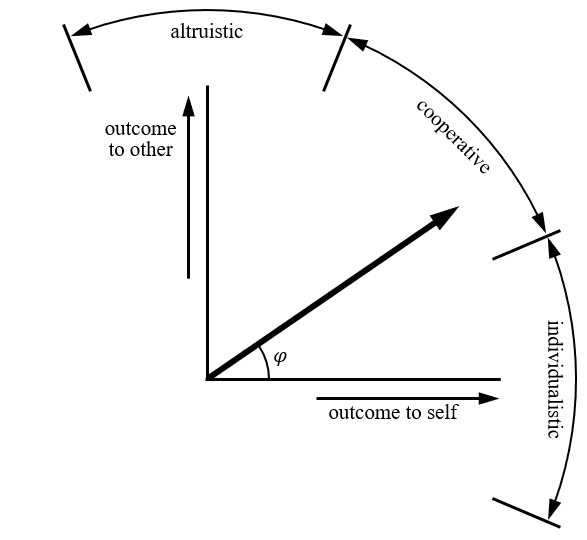}
    \caption{The range of SVO variation\cite{liebrand1988ring}.}
    \label{fig:svo ring}
\end{figure}

Generally, the SVO can be taken in the interval of $[-\pi,\pi]$, while some range of the SVO is not consistent with the general driving tasks. We restrict the value of SVO to $[0,\pi/2]$. The SVO can be shown by a part of the circle in Fig. \ref{fig:svo ring}. For example, when $\varphi=0$, it means an individualist who considers only the outcome to self; when $\varphi=\pi/4$, it means a cooperativist who considers both the outcome to self and others; when $\varphi=\pi/2$, it means an altruist who considers only with the outcome to others. To maximize the final utility, we need to determine $\varphi_1^*$ of the optimal SVO to satisfy:
\begin{equation}
    \varphi^* := \arg\min\limits_{\varphi} \left(U\cos(\varphi)+\hat{U}\sin(\varphi)\right),
\end{equation}
where $\varphi^*$ can maximize one's final utility that can be obtained after determining the respective outcome for self and others. This optimal SVO can be used to directly respond to the social interaction behavior of the vehicle in any current state. 

Bringing in the outcome of the vehicle in the interaction scenario, we can obtain the optimal SVO in the interaction scenario:
\begin{equation}
    \varphi^*= \arctan \left(\frac{\hat{U}}{U}\right) =\arctan\left(\frac{\hat{v}\hat{d}}{vd}\right).
\end{equation}

\subsubsection{Reward Function}
To achieve human-like social interaction behaviors, we can draw guidance from naturalistic driving data. To do so, we propose using the previously described SVO as an intermediate variable that reflects human interaction behaviors during the learning process driven by data. The value of SVO can be derived from the dataset.

\textbf{SVO reward}: The SVO responds to the allocation of outcomes between the two interacting parties by the ego vehicle in the current state, and the similarity to the SVO in the naturalistic driving dataset means the approximation to the real-world interaction logic, which guides the vehicle to achieve human-like interaction behavior by penalizing the error between the SVO of the vehicle and the SVO in the naturalistic driving dataset at the same time step:
\begin{equation}
    r_1 =
    \begin{cases}
        -\frac{\left|\varphi_{t+1,GT} - \varphi_{t+1}\right|}{0.5}, &\text{if}\ t+1 \leq T\\
        -0.5, &\text{else} 
    \end{cases},
\end{equation}
where $r_1$ represents the reward related to the SVO. $T$ is the total number of time steps of the trajectory. $\varphi_{t+1, GT}$ is the SVO sequence in dataset and $\varphi_{t+1}$ is the vehicle's SVO in next time step.

\textbf{Episode length reward}: Considering the interaction of two vehicles from the starting position to the point where one of the vehicles reaches the conflict point as a complete interaction, the episode length represents the time required to complete an interaction task. If the vehicle's behavior exactly replicates the human driver's behavior in the dataset, then the vehicle's episode length should be the same as the trajectory time length in the naturalistic driving dataset. We design the following penalty based on the error of episode length and trajectory time length:
\begin{equation}
    r_2 = -\frac{|l_{epsiode} - l_{GT}|}{l_{GT}},
\end{equation}
where $l_{episode}$ denotes the number of time steps in a complete interaction, and $l_{GT}$ denotes the number of steps in a complete interaction in the same interaction scenario in the dataset. The maximum value of $r_2$ is determined by the maximum allowable number of steps in the simulation environment (set here to twice the $l_{GT}$).

\textbf{Velocity reward}: According to the definition of SVO proposed above, the value of SVO does not change when the velocities of the two vehicles are scaled up or down in equal proportions, which requires additional evaluation related to the velocity to restrict vehicles. We can use the error of the episode length and trajectory time length to evaluate the velocity reward. If the episode length is larger than the trajectory time length, it means that the overall velocities of two vehicles are low, so we need to give it large penalty. Similarly, we need to give a large penalty for the high velocity. We use the sum of the velocities of the two vehicles at the next time step as an overall velocity indicator, weighted by the time length error to obtain the penalty of velocity:
\begin{equation}
    r_3 = -\frac{(v_{t+1} +\hat{v}_{t+1})}{20} \times \frac{(l_{episode} - l_{GT})}{l_{GT}},
\end{equation}
where $v_{t+1}$ represents the velocity of the ego vehicle at the next time step, and $\hat{v}_{t+1}$ represents the velocity of the other vehicle at the next time step.

\textbf{Safety reward}: To ensure basic safety performance during vehicle interaction, we give a high penalty to the occurrence of each collision, which is used to eliminate the possibility of collision behavior to the greatest extent possible:
\begin{equation}
    r_4=\begin{cases}
         -10, &\text {if} \ collision\\
         0,   &\text{else}
    \end{cases}.
\end{equation}

Based on the above reward terms, we can obtain $r_1\in[0,1]$, $r_2\in[0,1]$, $r_3\in[-1,1]$ and $r_4\in\{-10,0\}$. We linearly combine each reward term to obtain the final reward function that guides the RL method in finding the optimal policy:
\begin{equation}
    r = \alpha_1r_1+\alpha_2r_2+\alpha_3r_3+\alpha_4r_4,
\end{equation}
where $\alpha_i$ represents the weights of different reward terms, which are set as $\alpha_1=-1,\alpha_2 = -2,\alpha_3=-1,\alpha_4=1$ after the parameter calibration.

\subsection{Human-like Interaction Decision Model Considering Social Interaction}
\subsubsection{Vanilla Soft Actor-Critic Method} 

In general, RL methods can be divided into value-based and policy-based methods. Value-based methods acquire a policy by learning the value function, and there is no displayed policy in the learning process, whereas the value-based approach cannot handle continuous action space and high action dimension scenarios. In contrast, the policy-based approach learns a displayed target policy directly, whereas the policy-based approach is not efficient and has a high deviation. To combine the above advantages and overcome the disadvantages, the actor-critic architecture is proposed in many widely-used RL methods such as Deep Deterministic Policy Gradient (DDPG) \cite{lillicrap2015DDPG}, Proximal Policy Optimization (PPO) \cite{schulman2017PPO}, Soft Actor-Critic (SAC) \cite{haarnoja2018SAC}.

\begin{figure*}[tbp]
\begin{equation}
L_Q(\phi)=\mathbb{E}_{(s_t,a_t,r_t,s_{t+1})\sim\mathcal{D},a_{t+1}\sim\pi_{\theta(\cdot|s_{t+1})}}\left[\frac{1}{2}(Q_{\phi}(s_t,a_t)-(r_t+\gamma(\min\limits_{j=1,2}Q_{\overline\phi_j}(s_{t+1},a_{t+1})-\alpha\log\pi(a_{t+1}|s_{t+1}))))^2\right]
\end{equation}
\end{figure*}

\begin{figure*}[tbp]
\begin{equation}    L_\pi(\theta)=\mathbb{E}_{s_t\sim\mathcal{D},\epsilon_t\sim\mathcal{N}}\left[\alpha\log(\pi_\theta(f_\theta(\epsilon_t;s_t)|s_t))-\min\limits_{j=1,2}Q_{\phi_j}(s_t,f_\theta(\epsilon_t;s_t))\right]
\end{equation}
\end{figure*}

The SAC is a model-free off-policy RL method based on actor-critic architecture. The actor outputs actions through the policy net, and the critic evaluates the output of the action by the Q-value network and updates the policy net. In this paper, we enhance the SAC to adapt to solving the concerned problem. In the general RL approach, the learning target is to maximize the cumulative reward:
\begin{equation}
    \pi^* = \arg\max\limits_{\pi}\mathbb{E}_\pi\left[\sum\limits_t r(s_t,a_t)\right],
\end{equation}
where $r(s_t,a_t)$ denotes the reward value obtained after performing action $a_t$ in state $s_t$.

To increase the randomness of the policy, the SAC adds an additional regular term of entropy to the objective, which makes the learning objective become:
\begin{equation}    \pi^*=\arg\max\limits_\pi\mathbb{E}_\pi\left[\sum\limits_t {r(s_t,a_t)}+\alpha H(\pi(\cdot|s_t))\right],
\end{equation}
where $H(\pi(\cdot|s_t))$ denotes the randomness of the policy $\pi$ in state $s_t$, $\alpha$ is the regularization coefficient of entropy, which is used to control the importance of entropy. The larger the value of $\alpha$, the higher the importance of entropy and the more the exploration of the agent, which will help to find the optimal strategy and can reduce the possibility of falling into a local optimum.

There are five networks in the SAC, including one policy network, two Q-value networks, and two target Q-value networks. The update of the Q-value network refers to the idea of DDQN \cite{van2016deep}, where the smaller Q-value is selected to participate in the update in two Q-value networks for the problem of the overestimation of Q-values. The loss function of the Q-value network is defined as in Eq. (13). $\mathcal{D}$ denotes the replay buffer which stores the tuple of the state transition process in off-policy RL algorithm, $(s_t,a_t,r_t,s_{t+1})$ is a mini-batch of tuples samples randomly form the replay buffer.

The update of the policy network can be viewed as the process of minimizing the KL-divergence, and since the policy output is Gaussian distributed mean and variance, the policy function needs to be made derivable by reparameterizing the sampling, and the loss function of the policy network is expressed in Eq. (14). $\epsilon_t$ is a random noise and $f_\theta(\epsilon_t;s_t)$ is the action after reparameter sampling.

To ensure the stability of the training results, the regularization coefficient of entropy needs to be adjusted. The SAC uses the method of automatic adjustment to set the loss function of regularization coefficient $\alpha$:
\begin{equation}    L(\alpha)=\mathbb{E}_{s_t\sim\mathcal{D},a_t\sim\pi(\cdot|s_t)}\left[-\alpha\log\pi(a_t|s_t)-\alpha\mathcal{H}_0\right],
\end{equation}
where $\mathcal{H}_0$ denotes the target entropy set artificially, when the entropy of current policy is larger than $\mathcal{H}_0$, the loss function will decrease the value of $\alpha$ to make more focus on value enhancement; when the entropy of current policy is smaller than $\mathcal{H}_0$, the loss function will increase the value of $\alpha$, raising the level of exploration of the policy.

\subsubsection{Enhancement by the Episode-replay Reward}

The goal of RL methods is to find an optimal policy to maximize the reward function. In this work, we want to find a policy that the social interaction behavior of vehicles is similar to the naturalistic driving dataset to obtain human-like decision-making at unsignalized intersections. Unlike regular RL methods, some information related to the interaction needs to be available after the entire interaction process is complete (e.g., interaction duration, conflict points, etc.). For the early interaction phase of the vehicle, the rewards associated with the interaction information are extremely delayed. However, for reproducing or learning the whole interaction process, the more early deviations may have a greater impact on the whole process. Therefore, we introduce the concept of episode-replay reward, where the final interaction information is obtained at the end of a complete interaction process. Then, the final interaction information will be fed back to each step of the whole process, and the reward function is recalculated for each time step.

\begin{figure*}[t]
    \centering
    \includegraphics[width=7in]{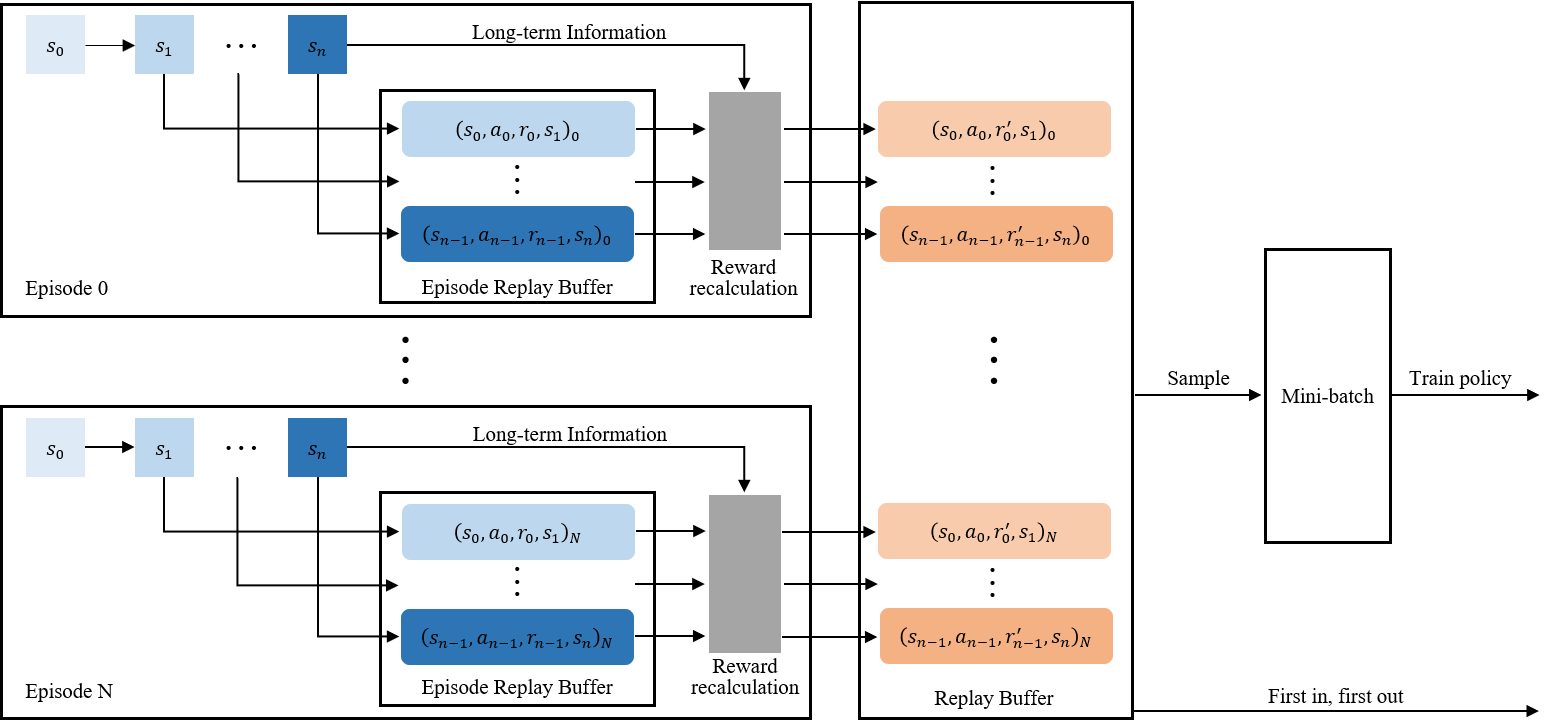}
    \caption{Episode-replay reward.}
    \label{fig:Episode replay reward}
\end{figure*}

The specific flow of the episode-replay reward is shown in Fig. \ref{fig:Episode replay reward}. In each episode, a separate episode-replay buffer is set up to store the state transfer tuple of the current episode. A new state transfer tuple is deposited into this episode-replay buffer for each interaction with the environment. When an episode ends, we can get long-term information about the whole episode. Combining this long-term information, we can recalculate the reward of each transfer tuple in the episode-replay buffer, and generate new state transfer tuples. Then, these new state transfer tuples will be deposited into the replay buffer of the entire RL framework. When the number of tuples stored in the replay buffer exceeds the minimum value, a mini-batch is randomly sampled from it for policy learning. Once the replay buffer reaches its maximum capacity, it will be updated with a first-in-first-out policy to remove previously-stored tuples.

The detailed process of the proposed Soft Actor-Critic with Episode-replay Reward (SACER) method is shown in Algorithm 1.

\begin{algorithm*}
	\caption{SACER: Soft Actor-Critic with Episode-replay Reward} 
	\begin{algorithmic}[1]
            \STATE Initialize replay buffer $\mathcal{D}$;
            \STATE Initialize $\phi_1, \phi_2, \theta$;
            \STATE Initialize target network with $\overline{\phi_1}\leftarrow\phi_1,\overline{\phi_2}\leftarrow\phi_2$;
		\FOR {each episode}
                \STATE Initialize episode replay buffer $\mathcal{D}_{episode}$;
		        \WHILE{not $done$}
		          \STATE Observe state $s_t$;
                    \STATE Select action $a_t$ from policy $\pi_{\theta}$ and step forward by Eq. (12);
                    \STATE Observe next state $s_{t+1}$, safety reward $r_{4,t}$ and done signal $done$;
                    \STATE Calculate SVO reward $r_{1,t}$ by Eq. (6);
                    \STATE Store $\left(s_t,a_t,r_{1,t},r_{4,t},s_{t+1},done\right)$ in $\mathcal{D}_{episode}$;
		        \ENDWHILE
                \STATE Get length of the episode $L_{episode}$;
                \FOR {each step in this episode}
                    \STATE Sample $\left(s_t,a_t,r_{1,t},r_{4,t},s_{t+1},done\right)$ sequentially from $\mathcal{D}_{episode}$;
                    \STATE Calculate episode length reward $r_{2,t}$ by Eq.(7);
                    \STATE Calculate velocity reward $r_{3,t}$ by Eq. (8);
                    \STATE Recalculate long-term reward $r_t$ by Eq. (10);
                    \STATE Store $\left(s_t,a_t,r_t,s_{t+1},done\right)$ in $\mathcal{D}$;
                    \STATE Sample a batch of transition $\left(s_t,a_t,r_t,s_{t+1},done\right)$ from $\mathcal{D}$ randomly;
                    \STATE Update Q-function by  $\phi_i\leftarrow\phi_i-\lambda_Q\nabla_{\phi_i}L_Q(\phi_i)\ for\ i \in \{1, 2\}$ in Eq. (13);
                    \STATE Update the policy by $\theta \leftarrow \theta-\lambda_{\pi}\nabla_{\theta}L_{\pi}(\theta)$ in Eq. (14);
                    \STATE Update the entropy regularization coefficient by $\alpha \leftarrow \alpha - \lambda \nabla_\alpha L(\alpha)$ in Eq. (15);
                    \STATE Soft update the target network by $\overline{\phi_i}\leftarrow\left(1-\tau\right)\overline{\phi_i} + \tau\phi_i\ for\ i \in \{1,2\}$;
                \ENDFOR
		\ENDFOR
	\end{algorithmic} 
\end{algorithm*}

\section{Experiment}
\subsection{Experiment Setup}

\begin{figure}
    \centering
    \includegraphics[width=8cm]{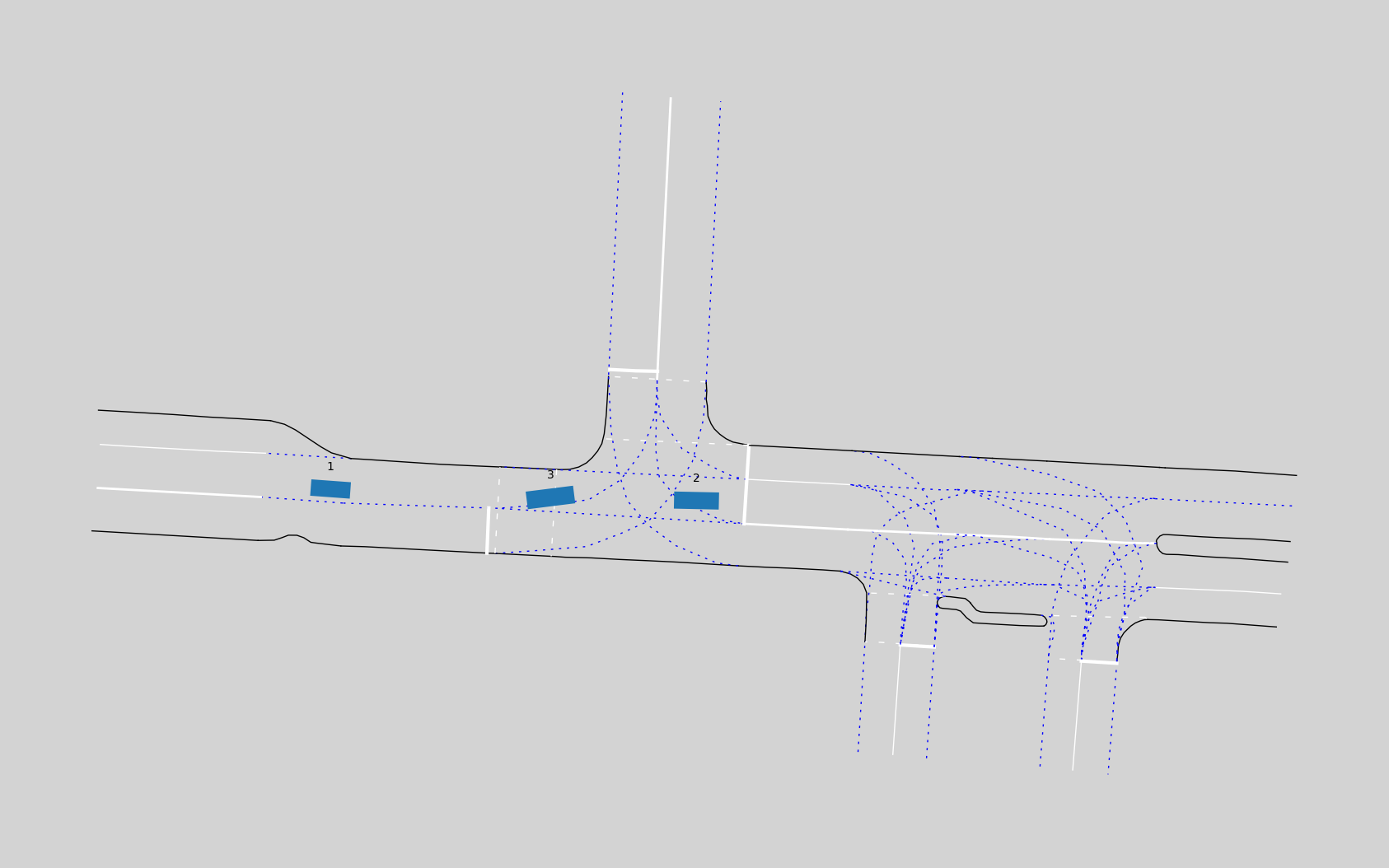}
    \caption{The unsignalized T-junction in the INTERACTION dataset.}
    \label{fig:Dataset}
\end{figure}

We select an unsignalized T-junction as the experimental scenario, $DR\_USA\_Intersection\_EP0$, from the INTERACTION dataset \cite{interactiondataset}, shown in Fig. \ref{fig:Dataset}. We obtain the real trajectories and specific information of the interacting vehicles from the dataset, and then we can extract many data samples with interactive processes under this scenario. In the experiment, 215 samples are used for training, and 17 samples are used for testing to evaluate the generalization performance of models.

The open source repository, highway-env \cite{highway-env}, is used to establish the environment for our task. Firstly, lanes were created based on the trajectory information of the pre-screened completed left-turning vehicles in the dataset and an opposing vehicle with which the interaction occurred. The initial positions and initial velocities of all vehicles were set based on the information from the dataset. The contours of all vehicles were set to be 5.0 m $\times$ 2.0 m. The frequencies of the simulation and policy update are set as 10 Hz. The maximum duration of each episode is defined as twice the duration of the complete interaction of that vehicle pair in the dataset. We specify the range of longitudinal acceleration values between -3.0 $m/s^2$ and 3.0 $m/s^2$. Each episode will run to generate vehicle trajectories until either vehicle reaches the interaction point. Any collision will cause the whole episode to end prematurely. In addition, exceeding the maximum time limit will also cause the episode to terminate.

\subsubsection{Models for Comparison}
In the experiment, the Behavior cloning (BC) and SAC with Episode-replay Reward from Velocity (SACER-V) are chosen as the comparison baselines to compare with our proposed method (SACER-SVO). The specific details of these methods are shown below:

\begin{itemize}
    \item \textbf{SACER-SVO} is our proposed method that utilizes the SACER architecture proposed above to introduce SVO in guiding the vehicle's policy during the learning process.
    \item \textbf{SACER-V} is to replace the SVO reward above with the deviation of the velocity of the vehicle from the velocity in the naturalistic driving dataset. In this way, we try to compare the advantages of SVO and velocity in guiding the RL-based method.
    \item \textbf{BC} is a supervised learning method that learns a deterministic policy by giving the states and corresponding action labels. The outputs of the BC method are the longitudinal accelerations of vehicles in interaction processes, which are consistent with the RL-based method.
\end{itemize}

\subsubsection{Implementation Details}

\begin{figure}[tbp]
    \centering
    \includegraphics[width=8cm]{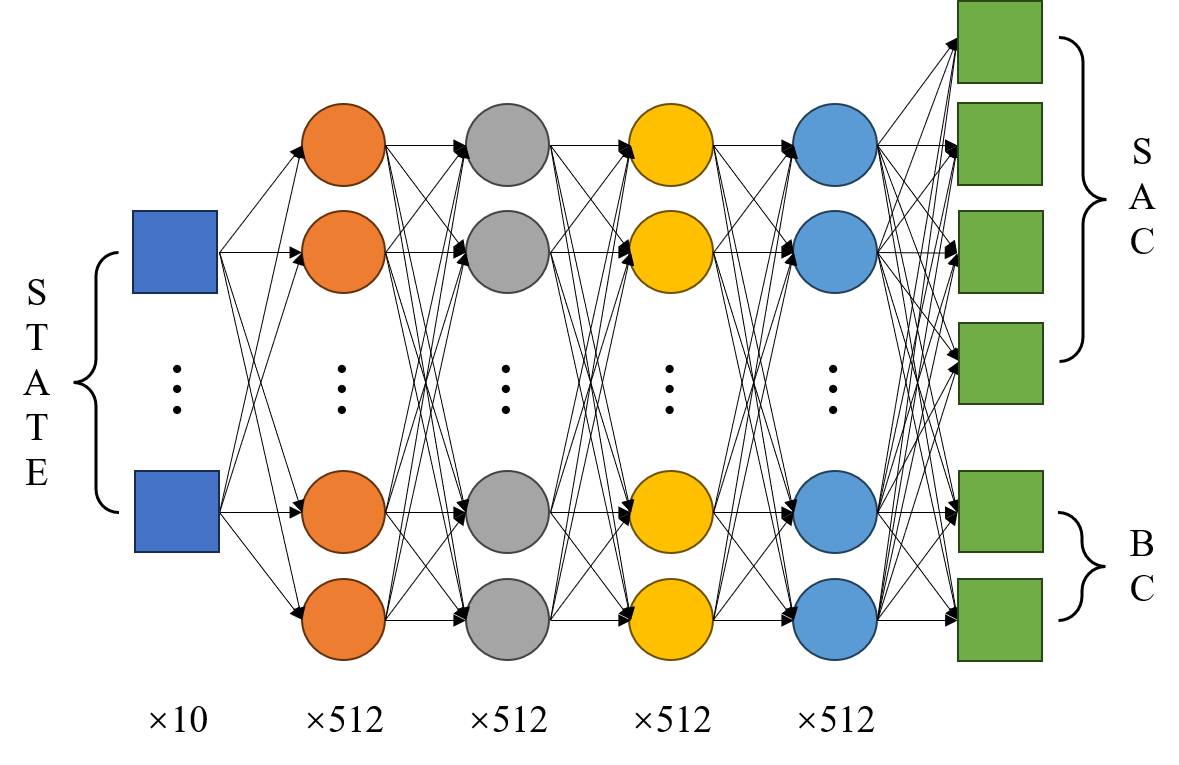}
    \caption{The structure of policy net.}
    \label{fig:policy net}
\end{figure}

All methods use a similar fully connected network structure, the specific structure of which is shown in Fig. \ref{fig:policy net}. We use a 4-layer fully connected network to implement the output of the policy, where the SAC-based method outputs the action distribution and the BC method outputs the deterministic policy. Other hyperparameters are shown in TABLE \ref{tab:hyperparameter}. Specifically in the proposed method, the strategy network outputs the mean and variance of each of the two actions, modeled as two independent Gaussian distributions, and the Q-value network outputs a deterministic value. The activation function of ReLU is used for all network layers except the output layer. We used the Adam optimizer to train the networks, and all networks were trained on PyTroch using an NVIDIA GTX 1660 Ti GPU. The proposed method requires approximately 19 hours of training time.

\begin{table}[tbp]
    \centering
    \caption{Parametric Notations and Settings.}
    \setlength{\tabcolsep}{0.8mm}{}
    \begin{tabular}{ccc}
    \toprule
    Parameter & Description & Setting \\
    \midrule
    $\lambda_\pi$ & Learning rate of policy network & 1e-4 \\
    $\lambda_Q$ & Learning rate of Q-value network & 1e-3 \\
    $\lambda$ & Learning rate of entropy regularization coefficient & 1e-4 \\
    $\gamma$ & Discount coefficient & 0.99 \\
    $\tau$ & Soft update parameter & 0.005 \\
    $\mathcal{H}_0$ & Target entropy & -2 \\
    / & Training episode times & 35000 \\
    / & Buffer size & 100000 \\
    / & Buffer minimal size & 1000 \\
    / & Batch size & 512 \\
    / & Simulation frequency & 10 \\
    / & Policy frequency & 10 \\
    \bottomrule
    \end{tabular}
    \label{tab:hyperparameter}
\end{table}

\subsection{Performance Evaluation}

\begin{figure}[tbp]
    \centering
    \includegraphics[width=8cm]{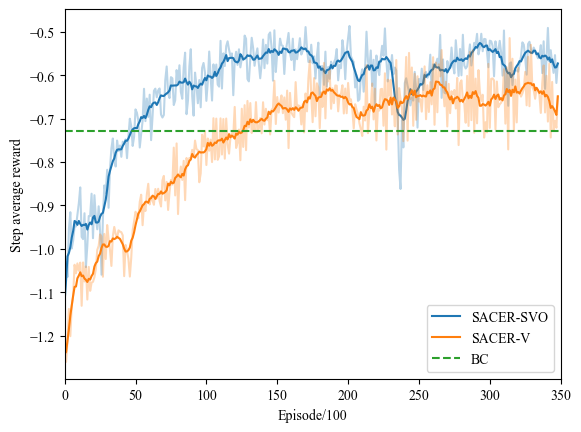}
    \caption{Reward curves for different methods in the training process.}
    \label{fig:Reward Curve}
\end{figure}

We compare the proposed model with the baseline method after training 35000 episodes. Since different pairs of interacting vehicles have different interaction durations, it is inaccurate to characterize the performance of the vehicle using the reward of the whole episode, and such reward curves are very volatile during the training process. We divide the total reward of the entire episode by the time steps in the episode to obtain the average step reward, which is used to characterize the iterations of the training process strategy. To compare different methods in the same metric, we recalculated the rewards of the SACER-V with the reward function of the SACER-SVO, and the specific training curves are shown in Fig. \ref{fig:Reward Curve}.

From Fig. \ref{fig:Reward Curve}, we can see that the average reward per step of our proposed SACER-SVO steadily increases during the training process and soon exceeds the performance of the BC method, and the training results stabilize after 25,000 episodes. Moreover, the SACER-SVO almost outperforms the SACER-V at almost all time steps.

Our goal is to obtain a more human-like strategy for the social interaction behavior of vehicles crossing the unsignalized intersection by learning from real-world datasets. As existing studies do not explore it in depth, there are no uniform parameters to measure social interaction performance. In this paper, we choose the following parameters for comparing the degree of human-like social interaction behavior of the strategies obtained from the models under different methods:

\textbf{Priority accuracy}:
Priority accuracy is the metric used to measure the result of the interaction behavior. Priority is considered accurate when the vehicles arrive at the conflict points in the same order as in the dataset, and vice versa with the wrong priority. 

\textbf{Episode length error}:
Episode length error represents the average of the difference between the time required by the vehicle during the interaction and the interaction completion time in the naturalistic driving dataset. It provides a comprehensive response to the degree of humanization of the interaction behavior, including the overall velocity and whether face-off occurs.

\textbf{Collision times}:
Collision times represent the total number of collisions under all cases and are used to directly measure the safety of the decision-making model.

\begin{table*}[htbp]
    \centering
    \caption{Quantitative Results for Different Models.}
    \setlength{\tabcolsep}{10mm}{}
    \begin{tabular}{ccccc}
    \toprule
    Model & Priority Accuracy & Episode Length Error & Collision Times \\
    \midrule
    BC-train & 72.09\% & 0.5540 & 1/215 \\
    SACER-V-train & 64.65\% & 0.2462 & 4/215 \\
    SACER-SVO-train & 81.86\% & 0.3835 & 1/215 \\
    BC-test & 64.71\% & 0.5759 & 0/17 \\
    SACER-V-test & 76.46\% & 0.3843 & 1/17 \\
    SACER-SVO-test & 94.12\% & 0.3078 & 0/17 \\
    \bottomrule
    \end{tabular}
    \label{tab:quantitative result}
\end{table*}

The quantitative results in TABLE \ref{tab:quantitative result} show that the SACER-based method has a significant advantage in priority accuracy and episode length error, which directly reflect the human-like characteristics of interaction behaviors. It is directly related to the fact that the BC method only learns through limited data. The cumulative error brought by the closed-loop simulation will continuously amplify the defects of the BC method and eventually lead to a poorer level of anthropomorphism.

The SACER-V method has the smallest episode length error in the training scenario because the velocity directly affects the length of the episodes. The SACER-V method guides the training of the vehicle by the real velocity in the dataset, thus achieving a more accurate limit on the episode length. The proposed SACER-SVO method achieves the optimal priority accuracy in both training and testing scenarios, and the episode length error also surpasses the SACER-V in the testing scenario, which proves that the method has obvious advantages in human-like interaction behavior decisions. It also demonstrates the advantage of using SVO as an intermediate variable to guide the vehicle's learning compared to directly using velocity to achieve human-like interaction behavior decisions.

\subsection{Case Study}

We select several specific cases in the test samples to compare the interaction behavior under different decision-making models. Each case is marked by a ternary array shaped as $(file\_id, left\_id, obs\_id)$.

\begin{figure*}[tbp]
\centering
\subfigure[Left-turning vehicle.]{
\includegraphics[width=0.37\textwidth]{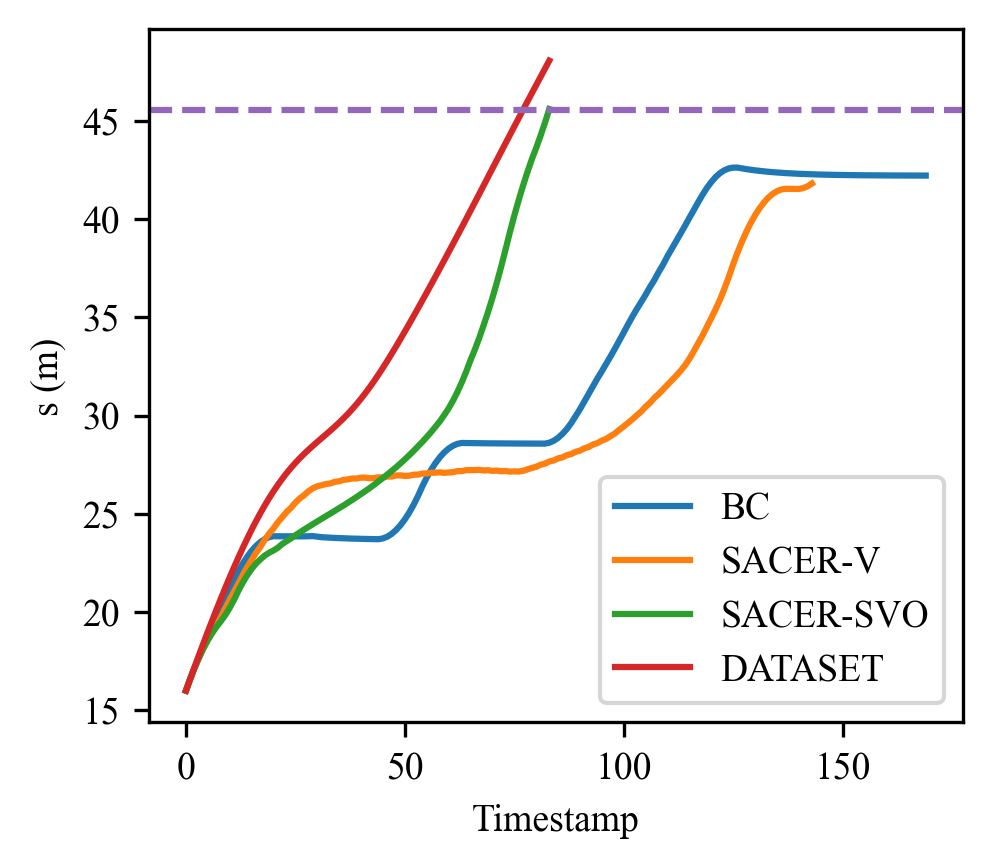}
\label{fig:s-t-1-left}}
\subfigure[Straight-going vehicle.]{
\includegraphics[width=0.37\textwidth]{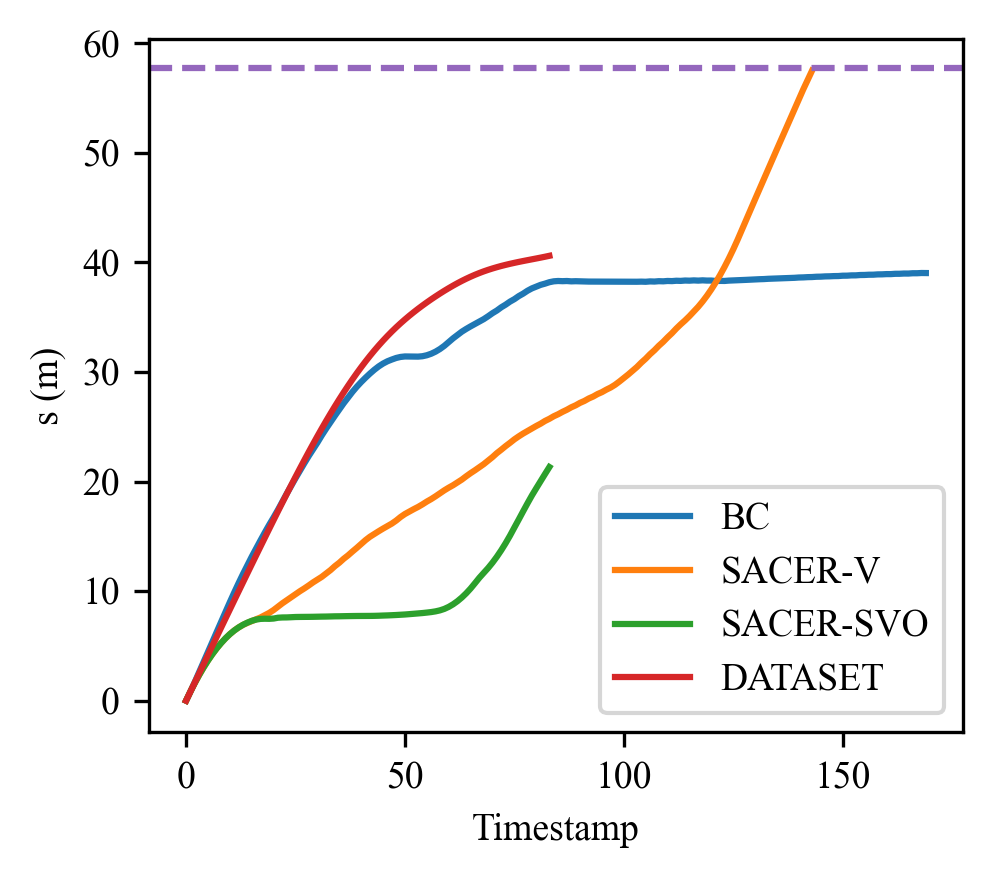}
\label{fig:s-t-1-obs}}
\caption{The s-t curves in case $(7, 25, 26)$.}
\label{fig:s-t-1}
\end{figure*}

In case $(7, 25, 26)$, a left-turning vehicle interacts with a straight-going vehicle in the opposite direction. The distances between the starting position of both vehicles and the conflict point are large. In the entire interaction process, the two vehicles slow down simultaneously before the straight-going vehicle eventually chooses to yield to the left-turning vehicle. 
To fully describe the interaction process of vehicles in this case under different models, the s-t curves of each vehicle can be seen in Fig. \ref{fig:s-t-1}. Four curves represent different decision-making models and the dashed line represents the location of the conflict point. The s-t curves of left-turning vehicles based on the SACER-SVO and dataset intersect with the dashed line, which means the vehicle under the SACER-SVO has the same priority as naturalistic driving data. However, the s-t curve of the straight-going vehicle based on the SACER-V intersects with the dashed line, which means the vehicle under the SACER-V has the wrong priority. In addition, neither vehicle based on the BC arrives at the conflict point, and the face-off situation (Both vehicles were unable to successfully pass the conflict point and forced each other to stop) occurs. Only the curve of the SACER-SVO stays close to that of the dataset in the horizontal coordinate direction. Fig. \ref{fig:s-t-1}(a) shows the s-t curves of the left-turning vehicle, and the s-t curve of the SACER-SVO shows the same trend as the s-t curve of the dataset. Fig. \ref{fig:s-t-1}(b) shows the s-t curves of the straight-going vehicle. Although the s-t curve of the BC remains almost the same as that of the dataset in the early stage, the vehicle chooses to stop without moving when approaching the conflict point, resulting in the whole interaction eventually not being completed successfully. In contrast, although the s-t curve of the SACER-SVO is somewhat different from that of the dataset, the whole interaction task is completed because the left-turning vehicle successfully reaches the conflict point. Unfortunately, all the s-t curves of the SACER-V differ greatly from the dataset, which means that the SACER-V cannot generate human-like decision results.

\begin{figure*}[htbp]
\centering
\subfigure[Left-turning vehicle.]{
\includegraphics[width=0.4\textwidth]{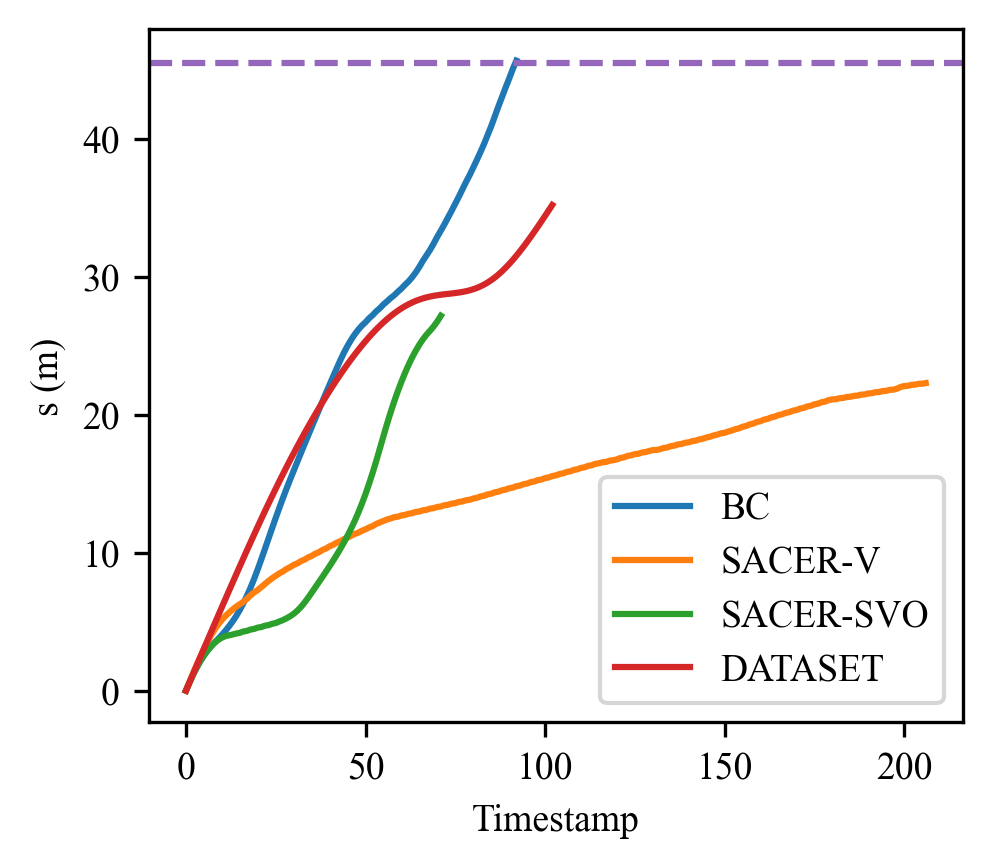}
\label{fig:s-t-2-left}}
\subfigure[Straight-going vehicle.]{
\includegraphics[width=0.4\textwidth]{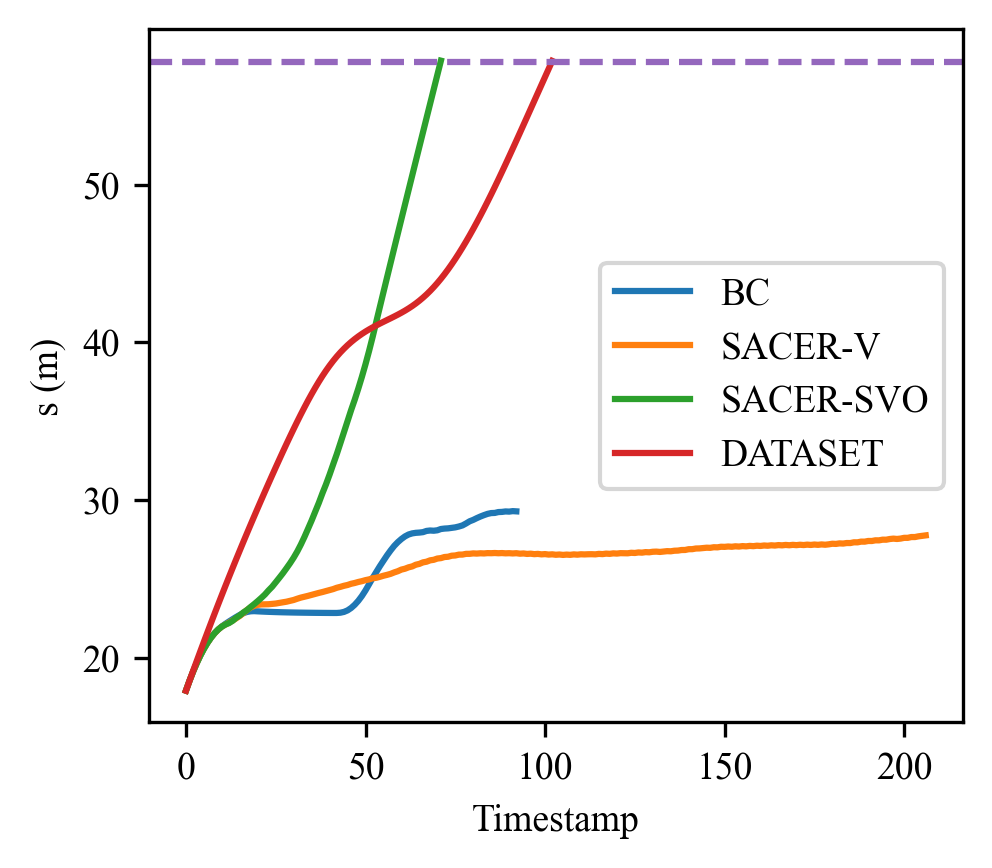}
\label{fig:s-t-2-obs}}
\caption{The s-t curves in case $(7, 49, 48)$.}
\label{fig:s-t-2}
\end{figure*}

In case $(7, 49, 48)$, a left-turning vehicle interacted with a straight-going vehicle in the opposite direction too, but the straight-going vehicles finally chose to pass the conflict point before the left-turning vehicles, which implemented a different interaction behavior from the previous case. The s-t curves of vehicles in this case can be seen in Fig. \ref{fig:s-t-2}. In the interactive scenario based on the SACER-SVO, straight-going vehicles first passed through the conflict point, maintaining consistent interaction priority with vehicles in the dataset. However, in the interactive scenario under the BC, the left-turning vehicle passed through the conflict point first, resulting in the wrong interaction priority to the dataset. In the interactive scenario based on the SACER-V, due to the low velocities, vehicles ultimately exceeded the maximum interaction time limit and did not complete the interaction task.

In case $(7, 57, 61)$, a left-turning vehicle interacted with a right-turning vehicle in the opposite direction. Eventually, the right-turning vehicle chose to yield to the left-turning vehicle. Unlike the previous cases, in this case, the trajectories of the two vehicles no longer cross but eventually merge into the same lane. In this case, the vehicles based on the SACER-SVO also maintained the same interaction priority as the dataset, where the left-turning vehicle passed the conflict point first. In contrast, the right-turning vehicle based on the SACER-V passed the conflict point first, resulting in a wrong interaction priority. In addition, the BC-based interacting vehicles appear to the face-off situation, with both vehicles stopping before the conflict point. SACER-SVO showed a consistent trend with the dataset on the s-t curves of both vehicles, reflecting a stronger level of anthropomorphism compared with other methods.

\begin{figure*}[htbp]
\centering
\subfigure[Left-turning vehicle.]{
\includegraphics[width=0.4\textwidth]{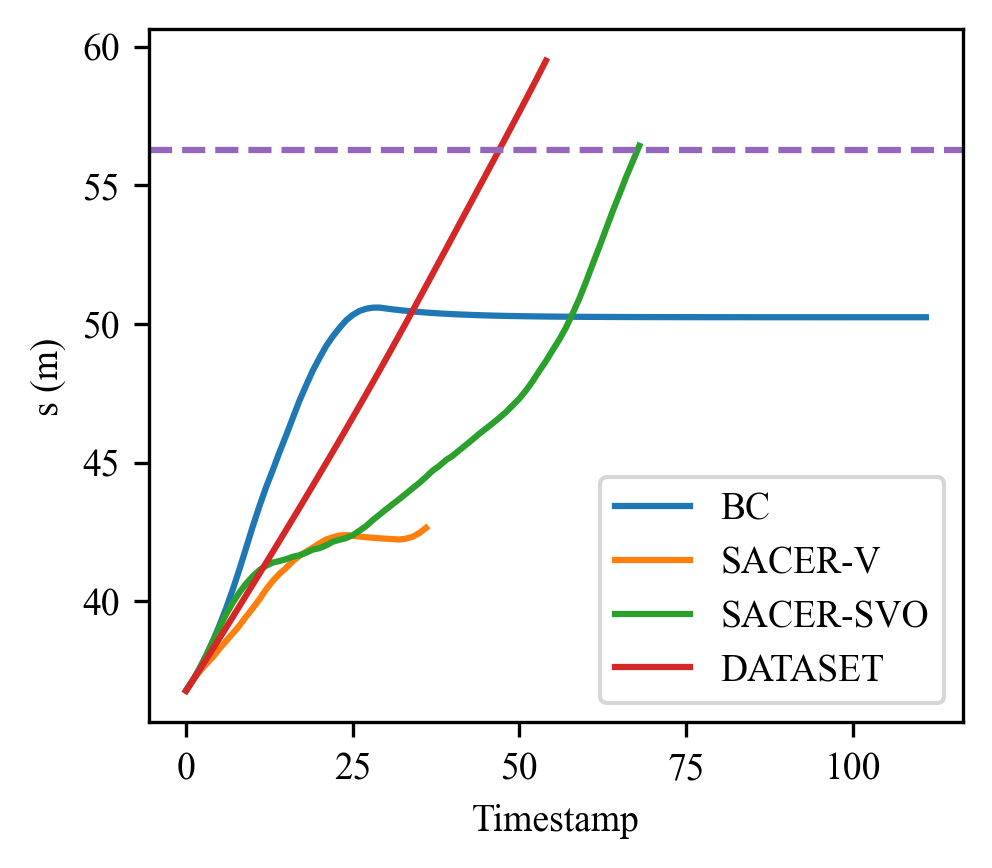}
\label{fig:s-t-3-left}}
\subfigure[Right-turning vehicle.]{
\includegraphics[width=0.4\textwidth]{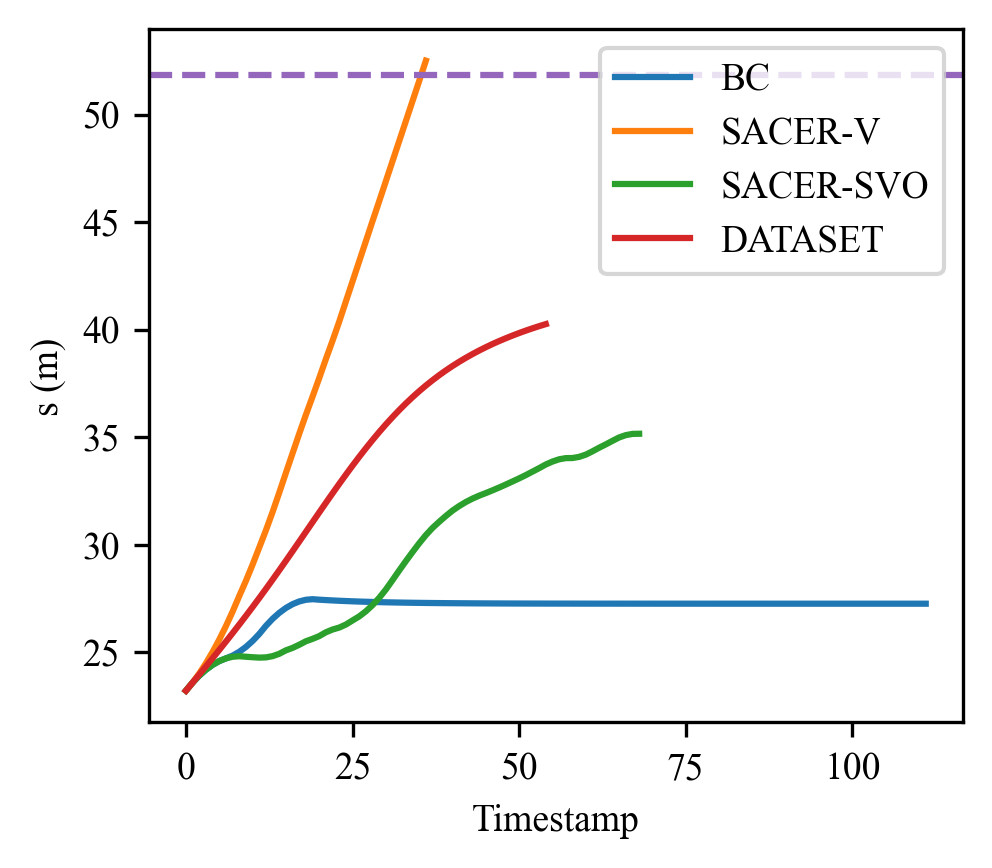}
\label{fig:s-t-3-obs}}
\caption{The s-t curve in case $(7, 57, 61)$.}
\label{fig:s-t-3}
\end{figure*}

\section{Conclusion}

This paper proposes a SACER-SVO model for human-like decision-making of vehicles at unsignalized intersections. The SVO is used to describe social interaction among multiple vehicles, which can be used as the intermediate variable to guide the vehicles' actions. Based on the naturalistic driving trajectories, the paper also provides quantitative results and case studies to demonstrate the effectiveness of the proposed method. The proposed model outperforms the baseline methods in priority accuracy and episode length error, which can accurately reflect human-like interaction behaviors. In future work, we will explore the coupling interaction process between more vehicles, and apply the SVO-guided decision-making methods to more complex scenarios.

\section*{Acknowledgments}
This work was supported by the Science and Technology Commission of Shanghai Municipality (Nos. 22YF1461400 and 22DZ1100102).

\balance

\bibliographystyle{IEEEtran}  
\bibliography{references}  
\end{document}